\documentclass[conference]{IEEEtran}

\IEEEoverridecommandlockouts
\usepackage[linesnumbered,ruled,vlined]{algorithm2e}
\usepackage{amsmath, amsfonts, bm}
\usepackage{amsmath,amssymb,amsfonts}
\usepackage{graphicx}
\usepackage{textcomp}
\usepackage{mathrsfs}
\usepackage{tikz}
\usepackage{pgfplots}
\usetikzlibrary{patterns}
\pgfplotsset{compat=1.18}
\usepackage{url}
\usepackage[sort&compress,numbers]{natbib}
\usepackage{hyperref}
\hypersetup{
    colorlinks=false,
    linkbordercolor={0 0 0},
    urlbordercolor={0 0 0},
    citebordercolor={0 0 0},
    pdfborderstyle={/S/U/W 0}
}
\usepackage{xcolor}
\usepackage[capitalize,noabbrev]{cleveref}

\def\BibTeX{{\rm B\kern-.05em{\sc i\kern-.025em b}\kern-.08em
    T\kern-.1667em\lower.7ex\hbox{E}\kern-.125emX}}
\begin{document}

\title{Sum of Costs Diffusion with Dynamic Guidance for Motion Planning\\
}

\author{
    \IEEEauthorblockN{Aysu Aylin Kaplan, Özgür Erkent}
    \IEEEauthorblockA{\textit{Computer Engineering Department, Hacettepe University}, Ankara, Turkey}
    \{aysukaplan, ozgurerkent\}@hacettepe.edu.tr
}

\maketitle

\begin{abstract}
The motion planning problem for robotic manipulation can be addressed through classical or deep learning approaches. Existing methods face significant challenges in generalizing to diverse settings. In this study, we present a method with high generalization capability that generates collision-free trajectories using diffusion models where the denoising process is guided by the gradient of the total collision cost. We are also presenting a dynamic approach for choosing start step of the gradient guidance. Experimental results demonstrate that guiding the diffusion model dynamically with the sum of collision costs offers more robust performance by overcoming the generalization issues faced by competing methods. The proposed model demonstrates its effectiveness by achieving the highest performance on diverse test settings in M$\pi$nets\ dataset among the compared methods. 
\end{abstract}

\section{Introduction and Related Works} 
Motion Planning problem can be defined as finding a continuous and collision free trajectory given a start and a goal point. This trajectory for a robotic manipulation task can be found using classical approaches~\cite{chomp,storm, lavalle2006planning}. Classical approaches can be divided as optimization or sampling based \cite{lavalle2006planning}. Optimization based approaches require extensive finetuning of cost functions and are highly dependent on initialization, whereas although being probabilistically complete, sampling based approaches often require significant computation time to find a solution. The motion planning problem for robotic manipulation task can also be solved using deep learning methods~\cite{motionplanning, motionpolicy}. These methods are highly data dependent and although they achieve high success rates in the settings where they were trained, they fail to generalize to new settings.

In recent years, diffusion models have demonstrated significant capability in image generation~\cite{dif2015, ddpm, podell2024sdxl}. Following their success in image generation tasks, recent literature have explored to solve motion planning problem using diffusion models \cite{gpd, edmp, potential-energy-dif, mpd, difseeder}. 
In one of the early studies, Motion Policy Diffusion used gradient of a single cost function to guide the generation process of the diffusion model~\cite{mpd}. EDMP showed that using only one type of a cost function for guidance is not enough for generalization~\cite{edmp}. They performed gradient guidance using different costs for each parallel denoising batch which resulted in improved performance. Depending on the success of EDMP, GPD used diffusion in polynomial parametric space of trajectories~\cite{gpd}. They employed a single cost for guiding the generation and performed stitching on the generated trajectories. GPD claimed that using polynomials as geometric priors improve computation time while maintaining collision avoidance performance. More recently, hierarchical diffusion frameworks have been extended from image synthesis to robotic control and planning \cite{ho2021cascadeddiffusionmodelshigh}. Subsequently, DPPO applied this structural logic to reinforcement learning by formulating a two-layer Diffusion Policy MDP \cite{ren2024diffusion}. Expanding this hierarchical approach to the motion planning, Cascaded Diffusion proposed a framework that unifies global and local prediction, feeding higher level model output into lower level diffusion model~\cite{cascaded}.


In this study, instead of using one cost function to guide the denoising process, we are calculating multiple costs and using the sum of these costs for the gradient guidance. By using the sum of costs, our approach achieves high success rates on both trained and new test settings while demonstrating better generalization than prior methods.


Another important difference in our gradient based guidance method is that, in contrast to prior methods~\cite{mpd, gpd, edmp}, which use intermediate noisy trajectory for the guidance, we compute the gradient using the predicted final trajectory. By utilizing the predicted final trajectory instead of the noisy trajectory for cost calculation, our approach achieves more accurate gradient based guidance performance.

The proposed method utilizes parallel denoising with gradient guidance, calculating the sum of costs via the final trajectory prediction. Furthermore, contrary to prior approaches that initiate gradient guidance at a fixed denoising step, our framework adaptively determines the start step for each problem by evaluating the weight uniformity across parallel processes weighted by their respective sum-of-costs.

Our contributions can be listed as follows:
\begin{itemize}
    \item We propose a guidance method that uses the gradient of the \textbf{sum-of-costs} during the denoising process of the diffusion model. 
    
    \item Unlike prior approaches that compute gradients on noisy intermediate states, we calculate the gradient on the \textbf{predicted final trajectory}.

    \item We introduce a \textbf{dynamic} gradient guidance initiation method based on the weight uniformity of the costs. 

    \item Performance has been improved compared to all other methods tested on the different settings of M$\pi$nets \cite{motionpolicy} dataset. 
\end{itemize}

\section{Approach}

\begin{algorithm}[t]
\caption{Sum-of-Costs Diffusion with Dynamic Guidance}
\label{alg:diffusion_guidance}

\KwIn{Trained denoising process $p_\theta$, diffusion timestep $T$, $K$ parallel processes, gradient guidance scale $w$, weight normalization scale $\lambda$.}
\KwOut{Trajectory $\boldsymbol{\tau}^{best}$.}

\BlankLine
\textbf{Initialization:}\;

Sample $K$ trajectories: $\boldsymbol{\tau}_T^{(1 \dots K)} \sim \mathcal{N}(\mathbf{0}, \mathbf{I})$\;

Set guidance active flag $G \leftarrow \text{False}$; $\tilde{U}_{T+1} \leftarrow 0$\;

\For{$t = T, T-1, \dots, 1$}{
    \ForEach{$k \in \{1, \dots, K\}$}{
        Predict final trajectory: $\hat{\boldsymbol{\tau}}_0^{(k)}$ (Eq.~\ref{eq:x0})
        
        Compute total cost: $J(\hat{\boldsymbol{\tau}}_0^{(k)}) $ (Eq.~\ref{eq:cost})\;
    }
    
    \If{$G$ is \text{False}}{
        Compute the weights $w_t$ (Eq.~\ref{eq:weight})\
        
        Compute the weight uniformity $U_t$ (Eq.~\ref{eq:diversity})\
        
        Compute $\tilde{U}_t = \gamma \tilde{U}_{t+1} + (1 - \gamma) U_{t}$\
        
        $\nabla_t = |\tilde{U}_t - \tilde{U}_{t+1}|$\
        
        \If{$\nabla_t < \epsilon$}{
            $G \leftarrow \text{True}$\;
        }
    }

    \ForEach{$k \in \{1, \dots, K\}$}{
        \If{$G$ is \text{True}}{
            Update via gradient: 
            
            $\tau_{t}^{(k)} \leftarrow \tau_{t}^{(k)} - w \nabla {J}(\hat{\tau}_0^{(k)})$\;
        }
        Sample denoised trajectory: $\boldsymbol{\tau}_{t-1}^{(k)} \sim p_\theta(\boldsymbol{\tau}_{t-1}^{(k)} \mid \boldsymbol{\tau}_t^{(k)})$\;
        
        Fix boundaries: $\mathbf{q}_0^{(k)} \leftarrow \mathbf{q}_{start}, \mathbf{q}_{L}^{(k)} \leftarrow \mathbf{q}_{goal}$\;
        
    }
}

\BlankLine
\textbf{Final Selection:}\;
Sort $\{\boldsymbol{\tau}_0^{(1 \dots K)}\}$ by ${J}(\boldsymbol{\tau}_0)$ and select the first collision-free trajectory as $\boldsymbol{\tau}^{best}$\;
\end{algorithm}

\textbf{Motion Planning Problem:}
Motion planning for robotic manipulation tasks aim to find a collision-free trajectory between a start configuration $\mathbf{q}_{s}$ and a goal configuration $\mathbf{q}_{g}$. A trajectory comprising $L$ waypoints is defined as $\tau = \{\mathbf{q}_0, \mathbf{q}_1, \dots, \mathbf{q}_L\}$. The objective is to obtain a trajectory $\tau$ that satisfies the boundary conditions $\mathbf{q}_0 = \mathbf{q}_{start}$ and $\mathbf{q}_L = \mathbf{q}_{goal}$ while avoiding collisions with workspace obstacles and self-collisions.

\textbf{Trajectory Generation using a Diffusion Model:}
Forward diffusion process is formally defined as a Markov chain that incrementally adds noise to a given data sample for $t$ timesteps, ultimately transforming the data into a standard normal distribution using $\beta_t$, the variance schedule at timestep $t$ and the identity matrix ${I}$:

\begin{equation}
q(\tau_t \mid \tau_{t-1}) = \mathcal{N}(\tau_t; \sqrt{1 - \beta_t} \tau_{t-1}, \beta_t {I})
\end{equation}

The proposed trajectory generation framework follows an iterative denoising scheme that transforms a Gaussian noise to a kinematically valid and collision-free path in the configuration space, as detailed in Algorithm \ref{alg:diffusion_guidance}. This process is formulated as a reverse diffusion sequence where, at each timestep $t$, the model predicts and removes noise to refine the trajectory. This iterative denoising process, parameterized by $\theta$, is defined as:
\begin{equation}
    p_\theta(\boldsymbol{\tau}_{t-1} \mid \boldsymbol{\tau}_t) = \mathcal{N}(\boldsymbol{\tau}_{t-1}; \boldsymbol{\mu}_\theta(\boldsymbol{\tau}_t, t), \boldsymbol{\Sigma}_\theta(\boldsymbol{\tau}_t, t))
\end{equation}
where $\boldsymbol{\mu}_\theta$ and $\boldsymbol{\Sigma}_\theta$ are the mean and covariance matrix of the Gaussian distribution computed given $\boldsymbol{\tau}_t$ and $t$.

The final trajectory estimate $\hat{\boldsymbol{\tau}}_0$ can be computed from the current noisy trajectory $\boldsymbol{\tau}_t$, where $\epsilon_\theta(\boldsymbol{\tau}_t)$ is the noise component predicted by the model~\cite{chung2022diffusion, efron2011tweedie}:
\begin{equation}
    \hat{\boldsymbol{\tau}}_0 = \frac{\boldsymbol{\tau}_t - \sqrt{1 - \alpha_t}\,\epsilon_\theta(\text{\boldmath$\tau$}_t)}{\sqrt{\alpha_t}}
    \label{eq:x0}
\end{equation}

\begin{figure}[t]
    \centering
    \includegraphics[width=1\linewidth, interpolate=true]{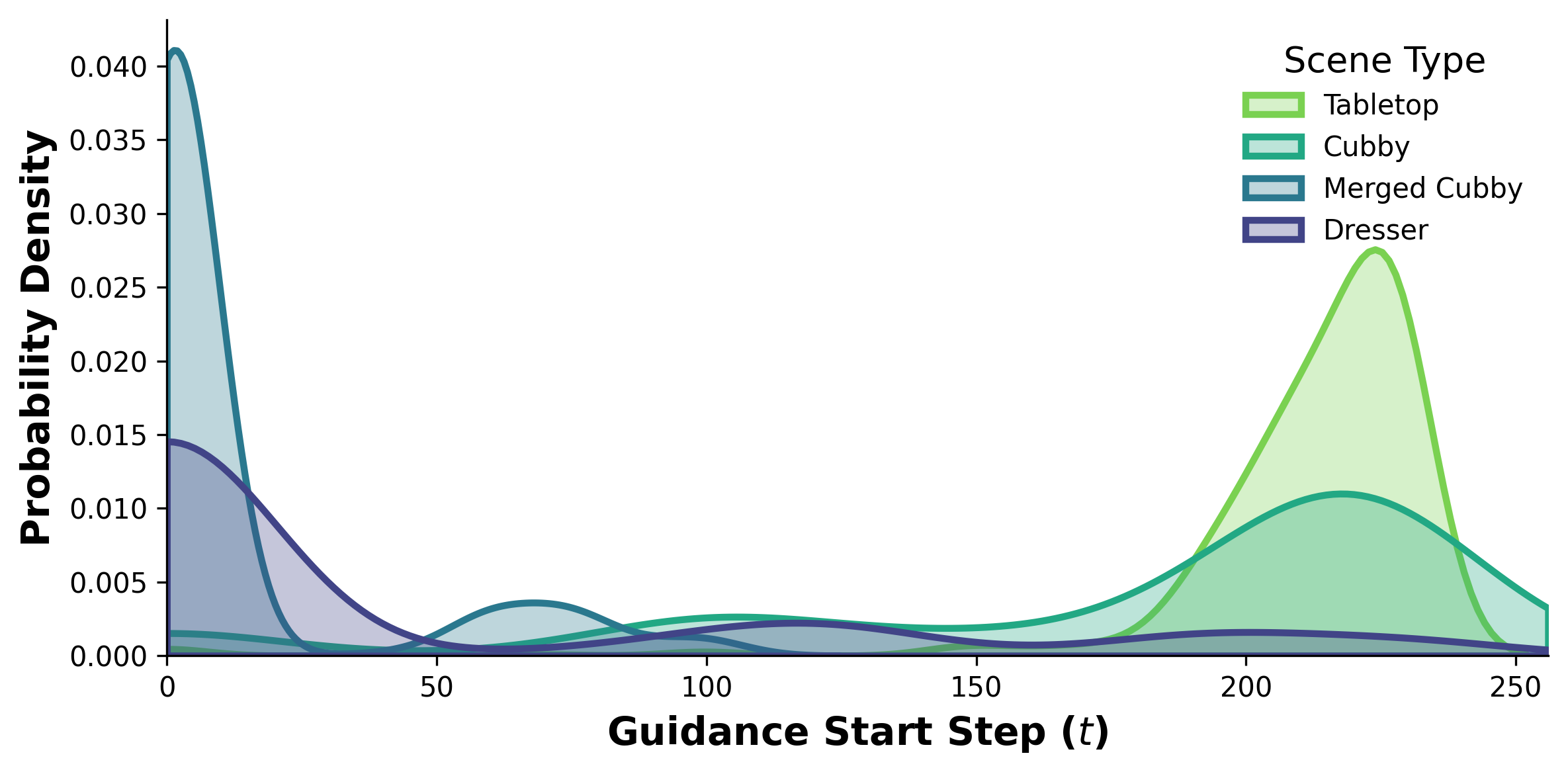}
    \vspace{-15pt}
    \caption{Probability Density of Gradient Guidance Initiation Timesteps Across Diverse Scenes on Both Test Set.}
    \label{fig:pdf}
\end{figure}

\textbf{Sum-of-Costs:}
We implement parallel denoising processes $k \in \{1, \dots, K\}$, and for each process at the  timestep $t$ of the denoising, we predict final trajectory $\hat{\boldsymbol{\tau}}_0^{(k)}$. The total cost function of each parallel process ${J}(\hat{\boldsymbol{\tau}}_0^{(k)})$ is computed using the predicted final trajectory $\hat{\boldsymbol{\tau}}_0^{(k)}$ as the sum of seven swept volume (SV) and five intersection volume (IV) costs \cite{edmp}:
\begin{equation}
    {J}(\hat{\boldsymbol{\tau}}_0^{(k)}) = \sum_{j=1}^{7} {J}_{SV, j}(\hat{\boldsymbol{\tau}}_0^{(k)}) + \sum_{l=1}^{5} {J}_{IV, l}(\hat{\boldsymbol{\tau}}_0^{(k)})
    \label{eq:cost}
\end{equation}

The intersection volume cost detects collisions by calculating penetration between robot and obstacle bounding boxes. The swept volume cost computes the volume swept between consecutive waypoints to prevent collisions between adjacent links. Each individual cost utilizes specific parameters, such as obstacle expansion and clearance distances. Summing these distinct cost functions enhances collision detection robustness by leveraging their different parameters. While a single cost may fail to register a specific contact due to its specific parameterization, the sum-of-costs approach ensures that collisions missed by one cost are captured by the other. By summing these individual volumes into one cost function, the total represents a more comprehensive approximation of the collisions. This sum-of-costs function guides the diffusion model's generation process, enabling the production of trajectories with enhanced generalization capability and safety guarantees.

\begin{figure}[t]
    \centering
    \includegraphics[width=0.8\linewidth, trim=2 8pt 0 0, clip]{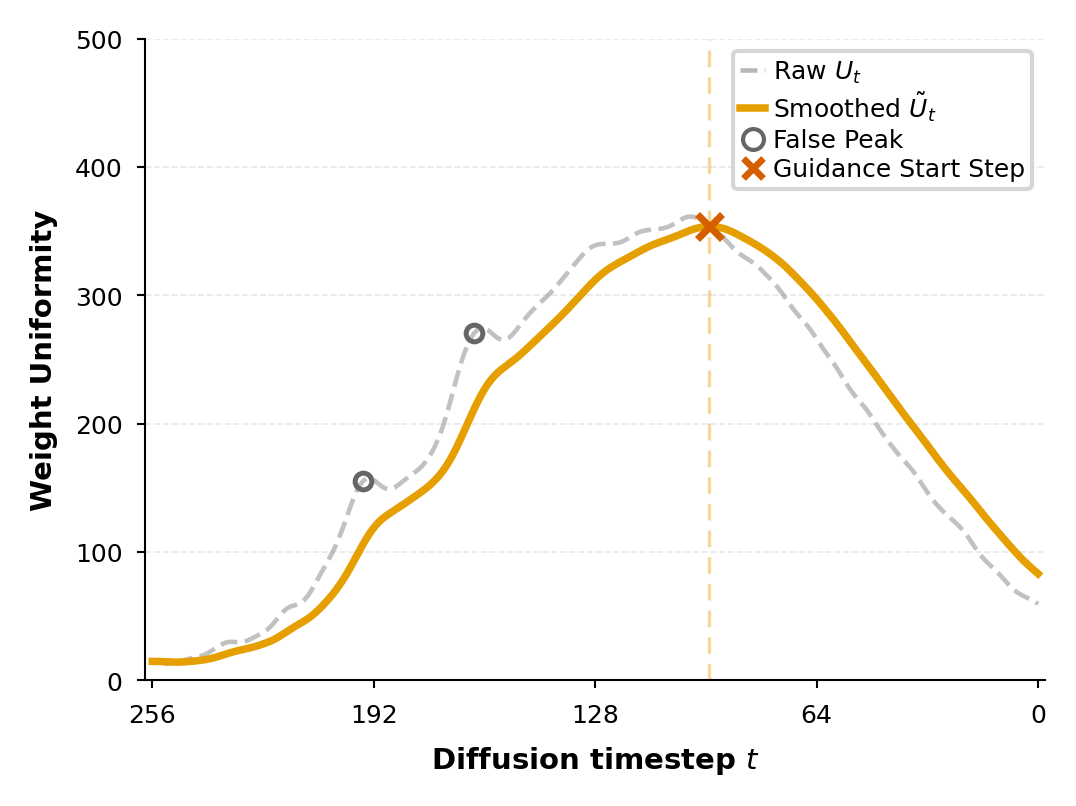}
    \vspace{-8pt} 
    \caption{Effect of EMA Smoothing on Guidance Activation.}
    \label{fig:ema}
\end{figure}

\textbf{Dynamic Start Step for Gradient Guidance:}
Fig.~\ref{fig:pdf} illustrates the empirical distribution of the initiation timestep across evaluated environments. To decide the start step for guidance, we first compute the normalized weights $w_t^{(k)}$ of each parallel process $k$, using softmax normalization with scale $\lambda$, where the parameter $\lambda$ acts as a temperature variable that regulates the sharpness of the distribution: 

\begin{equation}
    w_t^{(k)} = \frac{\exp(-{J}(\hat{\mathbf{\tau}}_0^{(k)}) / \lambda)}{\sum_{i=1}^{K} \exp(-{J}(\hat{\mathbf{\tau}}_0^{(i)}) / \lambda)}
\label{eq:weight}
\end{equation}

We then compute the weight uniformity (U), which quantifies how evenly the costs are distributed across parallel processes:
\begin{equation}
  \mathrm{U}_t = \Bigl[\sum_{i=1}^{K} (\hat w_t^i)^{2}\Bigr]^{-1}
\label{eq:diversity}
\end{equation}

Gradient guidance is dynamically initiated at the timestep $t$ where the smoothed weight uniformity $\tilde{U}_t$ reaches its global maximum as illustrated in Fig.~\ref{fig:ema}. This value is computed by applying an exponential moving average to the instantaneous weight uniformity $U_t$, defined as $\tilde{U}_t = \gamma \tilde{U}_{t+1} + (1 - \gamma) U_{t}$ where $\gamma \in [0, 1]$ is a smoothing coefficient. Mathematically, the initiation point is detected when the gradient $\nabla_t = |\tilde{U}_t - \tilde{U}_{t+1}|$ falls below a threshold $\epsilon$. This peak signifies the point of maximal cost uniformity across parallel processes, identifying the precise transition where stochastic noise is sufficiently reduced for accurate cost evaluation before the model collapses into specific local minima. Initiating guidance at this exact moment ensures that gradients steer the model based on valid structural predictions.

Once initiated, the denoising process is guided by the gradient of the cost function computed on the predicted final trajectory $\hat{\boldsymbol{\tau}}_0$, steering the model toward collision-free regions where the guidance scale $w$ controls the influence of the cost gradient on the denoising process:
\begin{equation}
    \boldsymbol{\tau}_t^{(k)} = \boldsymbol{\tau}_t^{(k)} - w \nabla {J}(\hat{\boldsymbol{\tau}}_0^{(k)})
\end{equation}

We propose calculating the cost gradient from the predicted clean trajectory $\hat{\boldsymbol{\tau}}_0$, in contrast to previous studies that compute guidance costs on the noisy trajectories $\boldsymbol{\tau}_t$. This formulation mitigates the effects of stochastic noise present in $\boldsymbol{\tau}_t$, which frequently yields unreliable cost estimations. Calculated on the denoised estimate $\hat{\boldsymbol{\tau}}_0$, the gradient represents the intended output more accurately. This alignment is critical because the primary objective is the generation of a collision free final trajectory; evaluating the cost on the predicted clean state ensures that guidance is directed toward the feasible trajectory space rather than being misdirected by intermediate noise present in $\boldsymbol{\tau}_t$. 


\textbf{Final Trajectory Selection:}
After denoising, the $K$ generated trajectories are sorted by their sum-of-costs in ascending order. The first collision-free trajectory in this sorted list is selected as the final trajectory $\boldsymbol{\tau}^{best}$.

\section{Experiments}
We are performing our experiments on 7 DoF Franka Emika Panda robot using PyBullet Physics simulator \cite{pybullet}.  

\subsection{Experimental Settings} 
\textbf{Dataset:} M$\pi$nets \cite{motionpolicy}, is a robotic motion planning dataset created to enable robotic manipulators to move through different environments without colliding with obstacles. M$\pi$nets has a hybrid training dataset consisting of 3.27 million collision-free trajectories on four type of scenes: tabletop, cubby, merged cubby and dresser. Hybrid training dataset is generated using a hybrid planner that combined AIT* \cite{ait*} and Geometric Fabrics \cite{g-fabrics}.
We evaluate our approach on the Hybrid, Both, and Global test sets of M$\pi$nets \cite{motionpolicy}. Each test set includes 1800 problems across the same four scene types as the training dataset. The Hybrid test set contains problems solvable only by the hybrid planner, while the Global test set consists of problems solvable exclusively by the global planner, AIT*~\cite{ait*}. The Both test set comprises problems solvable by both planners. Each problem consists of a start configuration, a goal position and obstacle configurations. 

\textbf{Model and Training Details:}  We train a temporal U-NET model similar to \cite{edmp} \cite{gpd} on the M$\pi$nets \cite{motionpolicy} Hybrid training dataset. We train the model for 7 hours on an RTX A6000 GPU for 2500 steps using a batch size of 16384 and a learning rate of 0.008 with SGD optimizer. We set the diffusion timestep to T=256 and the model outputs joint configurations for 50 waypoints along a trajectory. 

\textbf{Baselines:}  
In the experiments, we compare our approach against optimization based planners G.Fabrics \cite{g-fabrics}, CHOMP \cite{chomp} and STORM \cite{storm}, deep learning methods MPNets \cite{motionplanning} and M$\pi$nets \cite{motionpolicy}, diffusion based methods EDMP \cite{edmp}, GPD \cite{gpd} and Cascaded Diffusion \cite{cascaded}. For all deep learning and diffusion-based baselines, we report the performance of models trained on the Hybrid training set of M$\pi$nets \cite{motionpolicy}.

\textbf{Implementation Details:}  
We perform K parallel denoising process where we set K as 500. The selection of an ensemble size $K=500$ ensures that a large number of parallel processes explore the trajectory space, which facilitates the generation of more diverse and generalizable solutions. To implement gradient cost guidance, an empirically determined guidance scale $\omega = 0.1$ is utilized. Additionally, softmax cost normalization is performed using a scale factor $\lambda = 0.1$ which is selected to ensure that the weight distribution across parallel denoising processes is sufficiently sensitive to produce a distinct and identifiable peak using the uniformity metric $U_t$.

At each timestep, the denoising process may generate start and goal configurations that deviate from the problem specifications. Therefore, we enforce the boundary conditions by resetting the first and last waypoints of the trajectory to the target problem configurations after each denoising step, following ~\cite{mpd, edmp, gpd}.

Robot and obstacles are modeled as AABBs (Axis-Aligned Bounding Boxes) in the Pybullet simulation for cost calculation. After the diffusion model generation, a final collision check is performed using robot and obstacle meshes.
A problem is considered successful if the generated trajectory is collision-free, including self-collisions. Success rate is the percentage of successful problems relative to the total number of problems as defined in M$\pi$nets~\cite{motionpolicy}.

\subsection{Experimental Results}
\begin{table}[t!]
\caption{Comparison of Success Rates(\%). The best results are indicated in \textbf{bold}, and the second-best results are \underline{underlined}.}
\label{tab:method_comparison}
\centering
\begin{tabular}{|l|c|c|c|}
\hline
\textbf{Method} & \textbf{Global} & \textbf{Both} & \textbf{Hybrid} \\
\hline
G.Fabrics & 38.44 & 60.06 & 59.33  \\
Storm     & 50.22 & 76.00 & 74.50 \\
CHOMP     & 26.67 & 32.20 & 31.61 \\
\hline
MPNets    & 65.28 & 67.67 & 41.33 \\
M$\pi$Nets   & 75.78 & \underline{95.06} & 95.33 \\
\hline
EDMP      & 75.93  & 85.06 & 86.13 \\
GDP       & \underline{87.00} & 92.61 & 92.83 \\
Cascaded Diffusion  & 85.13 & - & \textbf{98.00} \\
\hline
Ours      & \textbf{94.22} & \textbf{96.88} & \underline{96.66} \\
\hline
\end{tabular}
\end{table}
The proposed approach demonstrates competitive performance across the evaluated benchmarks as presented in Table \ref{tab:method_comparison}. Specifically, the method achieves the highest results on the Global and Both datasets among the tested baselines. In the Hybrid dataset, our performance is comparable to Cascaded Diffusion \cite{cascaded}, by a 1.34\% margin. This slight performance gap can be attributed to the hierarchical architecture of the cascaded approach, which sequentially feeds the output of a global model as a conditioned input into a local diffusion model. This multi-stage approach provides a more informed starting point compared to our method, which generates trajectories starting from Gaussian noise. 
Furthermore, while Cascaded Diffusion incorporates an explicit plan refinement stage on the final output, which likely contributes to its slightly higher success rate in the hybrid test set, our approach eliminates the need for such post-processing. Notably, unlike methods that rely on trajectory stitching \cite{cascaded, gpd} or trajectory optimization \cite{difseeder}, our framework generates valid paths directly through the guided denoising process.

The proposed method demonstrates strong generalization capabilities; while trained on the Hybrid dataset, it achieves its highest performance on the Both dataset. These results suggest that integrating sum-of-costs guidance with a dynamic start step enhances performance by adaptively initiating cost steering. 

\subsection{Ablations}

\textbf{Ablation on Dynamic Start Step: }
In Table \ref{tab:ablation_start_step}, we explored the effect of the start step to start gradient guidance on the success rate on the Both test set. We compared the fixed timestep guidance baseline and our proposed dynamic gradient guidance method.
The results in Table \ref{tab:ablation_start_step} indicate that initiating guidance at earlier stages of the denoising process (e.g., T=256) leads to higher success rates compared to late stage initiation (T=32), with performance increasing from 90.88\% to 96.22\%. This trend suggests that earlier intervention provides more iterations for cost guidance to effectively influence the trajectory before it reaches final convergence. While this trend suggests that early intervention provides more iterations for cost guidance to effectively influence the trajectory before convergence, our dynamic initiation method achieves the highest overall performance. As illustrated in Fig. 1, the dynamically selected start step varies significantly across scene types, demonstrating that adaptively determining the guidance start step for each problem optimizes the transition from generative sampling to cost guidance better than using a fixed start step.

\begin{table}[!t]
\caption{Ablation of gradient guidance start step on success rate (\%).}
\label{tab:ablation_start_step}
\centering
\begin{tabular}{|l|c|c|c|c|c|}
\hline
\textbf{Start Step} & \textbf{32} & \textbf{64} & \textbf{128} & \textbf{256} & \textbf{Dynamic(Ours)} \\ 
\hline
\textbf{Success Rate} & 90.88 & 94.33 & 95.44 & 96.22 & \textbf{96.88}\\ 
\hline
\end{tabular}
\end{table}

\begin{table}[!t]
\caption{Ablation study on gradient guidance using predicted trajectories, showing success rates (\%).}
\label{tab:ablation_pred}
\centering
\begin{tabular}{|l|c|c|c|c|c|}
\hline
\textbf{Guidance Trajectory} & \textbf{Global} & \textbf{Both} & \textbf{Hybrid} \\ 
\hline
Current Trajectory(${\tau}_t$) & 89.44 & 91.44 & 91.94 \\ 
\textbf{Predicted Trajectory($\hat{\tau}_0$)} & \textbf{94.22} & \textbf{96.88} & \textbf{96.66} \\ 

\hline
\end{tabular}
\end{table}

\textbf{Ablation on Gradient Guidance on the Predicted Trajectory:}
We compared the effect of the chosen trajectory for the cost calculation on the gradient guidance in Table \ref{tab:ablation_pred}. Calculating gradient guidance using the predicted final trajectory rather than the noisy intermediate trajectory yielded an approximately 5\% increase in success rate across all test sets. This improvement suggests that the predicted trajectory provides a more accurate representation for cost evaluation.

\section{Conclusion}
In this study, a trajectory generation approach is presented that steers the denoising process of a diffusion model by utilizing sum-of-costs guidance. By integrating a diffusion prior with an adaptive guidance start step, the method seeks to generate valid paths directly within the denoising phase.
Experimental results demonstrate strong generalization; although trained on a Hybrid dataset, our model outperforms all baselines on the Global and Both sets with the dynamic start mechanism. Ablation studies confirm that our dynamic guidance approach is more effective than starting guidance at a fixed start step using the predicted final trajectory for the gradient guidance.

\bibliographystyle{IEEEtran} 
\bibliography{ref}

\end{document}